**Assessing Language Disorders using Artificial Intelligence: a Paradigm Shift**

**Charalambos Themistocleous[1], Kyrana Tsapkini[2], Dimitrios Kokkinakis[3]**

[1]Department of Special Needs Education, University of Oslo, Oslo Norway

[2]Department of Neurology, Johns Hopkins University, Baltimore, MD, USA

[3]Department of Swedish, University of Gothenburg, Gothenburg, Sweden

## Author Note

Data collection and preliminary analysis were sponsored by the University of Oslo, Department of Special Needs Education. We have no conflicts of interest to disclose.

Correspondence concerning this article should be addressed to Charalambos K. Themistocleous, Department of Special Needs Education, Helga Engs hus 4.etg Sem Sælands vei 7 0371 OSLO.

Email: charalampos.themistokleous@isp.uio.no

**Abstract**

Speech, language, and communication deficits are present in most neurodegenerative syndromes. They enable the early detection, diagnosis, treatment planning, and monitoring of neurocognitive disease progression as part of traditional neurological assessment. Nevertheless, standard speech and language evaluation is time-consuming and resource-intensive for clinicians. We argue that using machine learning methodologies, natural language processing, and modern artificial intelligence (AI) for Language Assessment is an improvement over conventional manual assessment. Using these methodologies, Computational Language Assessment (CLA) accomplishes three goals: (i) provides a neuro-cognitive evaluation of speech, language, and communication in elderly and high-risk individuals for dementia; (ii) facilitates the diagnosis, prognosis, and therapy efficacy in at-risk and language-impaired populations; and (iii) allows easier extensibility to assess patients from a wide range of languages. By employing AI models, CLA may inform neurocognitive theory on the relationship between language symptoms and their neural bases. Finally, it signals a paradigm shift by significantly advancing our ability to optimize the prevention and treatment of elderly individuals with communication disorders, allowing them to age gracefully with social engagement.

**Assessing Language Disorders using Artificial Intelligence: A Paradigm Shift**

## 1 Introduction

Language impairments are evidenced in patients with Mild Cognitive Impairment (MCI), Primary Progressive Aphasia (PPA), Alzheimer's (AD), and Parkinson's Disease (PD). Patients with AD are characterized by a progressive deterioration in cognitive domains such as memory, executive functions, and language, which become more severe as the disease progresses. However, language impairments can appear before the clinical manifestation of AD and are already evidenced in patients with Mild Cognitive Impairment (MCI), both in the amnestic and non-amnestic variants (Petersen et al., 1999). Therefore, detecting language impairments early in MCI is critical to provide treatments that can prevent symptom progression. Speech, language, and communication symptoms vary in patients with neurodegenerative disorders as this may depend on the affected brain areas, especially in the left hemisphere. Symptom variation is especially evidenced in patients with Primary Progressive Aphasia (PPA), a progressive neurological condition primarily affecting speech and language (Gorno-Tempini et al., 2011; Mesulam, 1982, 2001). Specifically, patients with inferior frontal damage are characterized by agrammatism and speech apraxia as their primary symptoms and are subtyped in the nonfluent/agrammatic PPA variant. Patients with temporal lobe damage are characterized by semantic impairments, such as naming impairments, and are subtyped in the semantic variant PPA (svPPA). Finally, patients with temporal and parietal lobe damage are characterized by phonological errors, repetition, and naming deficits and are subtyped in the logopenic variant PPA.

Assessing language function early can inform clinical decisions concerning remediation and compensation of language functioning, the effects and progress of atrophy on language, condition prognosis, and diagnosis (Strauss et al., 2006). The speech and language assessment aims to determine the language functioning in patients with speech and language impairments and identify deterioration of language functioning (Georgia Angelopoulou et al., 2018; Battista et al., 2017; Beales et al., 2018; Fraser, Lundholm Fors, & Kokkinakis, 2019; Matias-Guiu et al., 2022; Thilakaratne et al., 2022; Thomas et al., 2018). Also, it can provide an account of language (re)learning due to therapy, as learning is required for proper recovery and compensation (Krakauer, 2006).

Based on recent findings from Computational Language Assessment (CLA) studies, a collective term we use to refer to both AI and Computational Assessment tools, we argue that CLA is a superior approach to manual language assessment as it can detect dementia symptoms early, monitor disease progression and evaluate treatment efficacy by offering a quick, easy, and quantifiable assessment of speech, language, and communication functioning (Themistocleous, 2023). They can become easily accessible to patients as apps on their phones or through dedicated websites, allowing direct access to neurocognitive assessments to patients and clinicians. In addition, it can provide access to populations without access to a clinic for assessment, such as patients with mobility impairments and disadvantaged socioeconomic and language backgrounds. Specifically, CLA implemented in a computational application can assist i. *teleconsultation* informing healthcare professionals on patients in remote locations about symptom progression; ii.

*telehomecare*, supporting clinicians and doctors that overview and provide patient care; iii. telemonitoring by providing evaluation data over time, and as such, it can work together with other *monitoring* devices, such as devices monitoring heart rate and blood pressure to provide a holistic picture of patients' condition; and iv. *teletherapy*, delivering speech-language pathology, audiology, and other therapy services at a distance (Dial et al., 2019). Furthermore, such applications can assist clinicians in assessing and scoring discourse from patients fast and with accuracy, allowing them to focus on things that matter, such as having more time to treat patients.

In the following, we discuss (i) the main complications of using manual diagnostic batteries for language assessment in dementia diagnosis (ii) discuss applications of CLA, i.e., studies of computation and automation for language assessment. (iii) Finally, we discuss the underlying technologies and provide the main branches of CLA.

## 2 Complications of using manual diagnostic batteries

Over the past 50 years, standardized neurocognitive examination tests and neurolinguistic batteries, such as the Boston Naming Test (BNT; Kaplan et al., 2001), Western Aphasia Battery-Revised (WAB-R Kertesz (2006)), Boston Diagnostic Aphasia Examination (BDAE; Goodglass & Kaplan, 1983), Psycholinguistic Assessment of Language Processing in Aphasia, (PALPA; Kay et al., 1992) have been serving as the primary tools for screening patients with speech, language, and communication deficits. They focus on expressive language, such as naming, word finding, fluency, grammar, and receptive language. Language tests complement other tests for perceptual motor function (e.g., visual perception and visuomotor skills), executive functioning (e.g., planning, organizing, decision making, and working memory), learning and memory (e.g., immediate memory, short-term memory, long-term memory), attention (processing speech and sustained, divided, and selective attention) (Lezak, 1995).

Manual language assessments have been helpful as a language assessment tool (Lezak, 1995). However, manual testing focuses on assessing narrow language domains, such as word and sentence repetition, fluency, and naming, but not language production and comprehension in communicative settings. Nevertheless, symptoms manifest in communication, such as the ability of an individual to recall known names of familiar persons or places to that specific individual, finding words, planning the speech in the conversation, and uttering a coherent and cohesive speech. The communicative aspects of language provide ecologically valid measures that better portray the language deficits of individuals (Cunningham & Haley, 2020; Hansen et al., 2022; Holt et al., 2017; Kong et al., 2018; Mueller et al., 2018; Salis & DeDe, 2022; Stark et al., 2022). Therefore, manual standardized assessments do not comprehensively represent speech, language, and communication functioning and cannot inform early dementia detection, diagnosis, prognosis, and disease progression monitoring.

Manual language assessments are that they are less sensitive to patients' idiosyncratic deficits and conditions (Beltrami et al., 2018; Drummond et al., 2015; Drummond et al., 2019). Patients may perform poorly on tests for reasons unrelated to the pathology, such as fatigue, lack of sleep, education, and socialization opportunities. Furthermore, manual methods vary in accuracy among studies (Beach et al., 2012). For example, Beach et al. (2012) the sensitivity of neuropathologic examination ranged from 70.9% to 87.3%, and the specificity ranged from 44.3% to 70.8%. That allows the comparison of outcomes produced from sessions conducted at different time points or from other clinicians.

Similarly, global cognition screening tools, such as the Mini-Mental State Exam (MMSE), vary in sensitivity and specificity (Folstein et al., 1975). MMSE contains twelve questions measuring language, memory, attention, and orientation in time and space. It is scored out of thirty (30) and takes approximately five to ten minutes to administer but may last longer in patients with cognitive impairment. For MCI, MMSE provides a cut-off of 27/28 with a sensitivity of around 76% and a specificity close to 75% (Damian et al., 2011). However, its sensitivity is poor for right-lateralized focal brain lesions and executive and visuospatial impairments (Strauss et al., 2006).

Patients may also use dialectal variations, such as African American Vernacular English (AAVE), without enough data and standard assessments (Barnes, 2022; Labov, 1977). Also, manual language assessment is usually conducted late as patients must visit a neurology clinic for manual testing when the symptoms have progressed substantially and are evident to patients and others (Strauss et al., 2006). However, early detection and precision in diagnosing a neurodegenerative disorder are critical to inform therapy and family planning. Also, manual language assessments are difficult to administer, as they can be done only by trained personnel, and they are time-consuming and stressful to patients, all of which impede repeated language screening and monitoring of disease progression. The grammatical analysis performed by a human depends on their training, skills, knowledge of grammar, time constraints, tiredness, and other factors that result in different scores depending on the person who conducts the analysis and the specific conditions.

Finally, a critical disadvantage of manual standardized batteries and tests is that they are limited to language communities with the resources to produce them. Moreover, it requires substantial expertise to be adapted to other language varieties with care for keeping on the psychometric properties. Consequently, the lack of manual evaluations discriminates the patients from smaller language varieties.

## 3 Computational Language Assessment

The shortcomings of manual standardized assessments can be addressed by employing Computational Language Assessment (CLA). This language assessment approach uses artificial intelligence, including machine learning, natural language processing, statistical modeling, and signal processing, to assess speech, language, and communication functioning. As such, we argue that CLA-based evaluation of patients has several advantages over manual approaches. First, CLA models are continuously improving and evolving through learning and retraining on data, thus becoming better at detecting language biomarkers of dementia. In contrast, the detection rate of language impairment cannot improve over time in language batteries, as there is no straightforward approach to updating manual batteries and tests.

Also, CLA can offer objective measures of language functioning that do not depend on examiners, expertise, the theoretical paradigms they adhere to, time constraints, and other factors and limitations, such as fatigue, lack of sleep, and use of a dialect such as AAVE. CLA can detect speech patterns and characteristics that are not visible to listeners by accessing gradient features with physical meaning (e.g., fundamental frequency and vowel formants) that correspond to known physical events (e.g., the vibration of the vocal folds and modification of oral cavity from the active articulator) and continuous measures that cannot easily be matched to a physical meaning, such as *i*-vectors and spectral moments (Dehak et al., 2011; DeMarco & Cox, 2013; Glembek et al., 2011; Hautamäki et al., 2013; Jiang et al., 2014; Lee et al., 2014; Scheffer et al., 2011; Themistocleous, 2014, 2016a, 2016b, 2017a, 2017b, 2017c; Themistocleous et al., 2022; Themistocleous et al., 2016b). This is critical as it can identify patterns clinicians cannot evaluate without computational tools and distinguish characteristics not identifiable from a multifactorial analysis of speech and language.

In contrast to manual assessments that focus on a single task at a time conducted in an artificial environment (cf. a picture description task, naming tasks, verbal, and semantic fluency tasks), CLA provides ecological speech, language, and communication measures by incorporating real-world information from conversation and discourse. The latter can reveal a significant effect of aphasia due to dementia (Stark et al., 2022; Stark Brielle et al., 2020). Discourse and conversation have been known to reflect the early effects of dementia, which suggests that discourse and conversation can provide early biomarkers of dementia.

Written language can also be analyzed using CLA. This analysis has been instrumental in providing early biomarkers of dementia (Toledo et al., 2014). For example, researchers analyzed discourse micro-structure and macro-structure (e.g., planning, text coherence, and cohesion) in a longitudinal computational study of autobiographies from Catholic sisters of the School Sisters of Notre Dame congregation (Danner et al., 2001) from the Nun Study of Aging and Alzheimer's Disease. They showed that lexical, syntactic measures and idea density could predict dementia over time. This project provided linguistic, social, physiological, and anatomical markers (the nuns had donated their brains for posthumous study) of dementia. Other researchers identified the decline in lexical and syntactic production in the works of British novelists Iris Murdoch and Agatha Christie (Le et al., 2011) and U.S. President Ronald Reagan (Berisha et al., 2014; Berisha et al., 2015) and argued that they predict language function deterioration due to dementia. For example, a comparative discourse analysis of talks produced by U.S. President Ronald Reagan, who was diagnosed with Alzheimer's disease in 1994, and President George Herbert Walker Bush, who had no known diagnosis of Alzheimer's disease, showed a significant reduction in the number of unique words for President Ronald Reagan over time and increase in conversational fillers and non-specific nouns over time.

CLA machine learning models can assess discourse productions automatically and provide scores of language functioning concerning (i) *discourse microstructure* (e.g., phonetics, phonology, morphology, syntax, semantics) and (ii) *macrostructure* (e.g., discourse planning, cohesion, and coherence (56)) (Beaugrande & Dressler, 1981; Grice, 1975). Therefore, these measures can be employed to assess the *linguistic competence of patients* (Chomsky, 1965), namely their knowledge of language grammar (e.g., phonetics, phonology, morphology, syntax, and semantics) and their *communicative competence* (Hymes, 1996; Murray et al., 2007), namely, how individuals use language in the appropriate social context, follow social norms, and connect with other individuals and settings.

Computational methods can facilitate the analysis of discourse micro- (e.g., sentence-level grammatical domains) and macrostructure (e.g., textual domains, topics, cohesion, and coherence). CLA can provide valuable information about an individual's speech, language, and communication functioning from a short discourse sample.

1. The acoustic and phonological analysis of speech production (e.g., speech articulation and prosody) evaluates the ability of individuals to perceive and produce speech sounds and prosody and determines the phonological representation of sounds. Speech impairments commonly exist in patients with apraxia of speech, aprosodia, and phonological impairments.
2. The morphological analysis informs about the ability of individuals to form words and select the grammatical information associated with word structure, such as verb tense and aspect and case in nouns and adjectives.
3. The syntactic analysis provides information about syntactic processing, informing about syntactic comprehension and production impairments. For example, language deficits in recalling names or actions can be measured by calculating parts of speech, such as verbs and nouns (Themistocleous, Webster, et al., 2020).
4. The semantic analysis determines the appropriate use of word and sentence meanings and coherence of speech productions linked to semantic understanding and production impairments, such as naming impairments.
5. Pragmatics and *theory of mind* are linked to the ability of individuals to infer the state of mind of others in the conversation and express and understand communicative functions such as politeness, humor, and irony, which can become impaired in patients with dementia.
6. *Conversation, such as the ability of an individual to exchange conversational turns.* CLA provides measures of social cognition by quantifying *talk-in-interaction* as a measure of individuals' capacity to follow the turn-taking dynamics and social conventions in conversations (Sacks et al., 1974; Schegloff, 1998; Schegloff et al., 1977). This involves the ability of individuals to follow the social norms of turn-taking interactions, such as providing cues (e.g., prosodic and syntactic cues) for turn-transitions and turn continuations, and knowing how to deal with adjacency pairs, e.g., question-answer pair, greeting-greeting pair, complaint-acceptance/denial, and invitation-acceptance/denial (Aijmer & Stenström, 2005; Akynova et al., 2014). Such ecological measures were previously only feasible in the clinic, as discourse and conversation analysis are laborious, requiring many hours of manual work conducted by an expert individual, usually a linguist. Furthermore, the manual analysis of discourse needs more consensus (Stark et al., 2022; Stark Brielle et al., 2020).

7. *Emotions* are linked to the ability of individuals to express and perceive emotions in conversation and impairments related to emotions, such as apathy and depression. For example, emotional expressions or words also manifest specific relationships to speech and language pathology, and prosodic cues, such as low fundamental frequency and reduced pitch accent variation, can indicate apathy and depression in speech production (Stockbridge et al., 2021; Wright et al., 2018).

These suggest that CLA is an improved approach over manual neurolinguistic assessment approaches and has the potential to revolutionize neurolinguistic assessment. Next, we discuss methodological approaches that aim to offer objective and accurate language assessments that are culturally and linguistically sensitive, thereby reducing potential biases and improving diagnostic accuracy.

## 4 Components of CLA: Assessment and Monitoring

CLA combines different theoretical and methodological approaches to provide an optimal model for assessing language, which includes Machine Learning models, Natural Language Processing, and acoustic analysis of speech productions. These components can be employed to understand the characteristics of speech, language, and communication and to provide specific applications of the assessment for diagnosis, prognosis, and treatment efficacy evaluation.

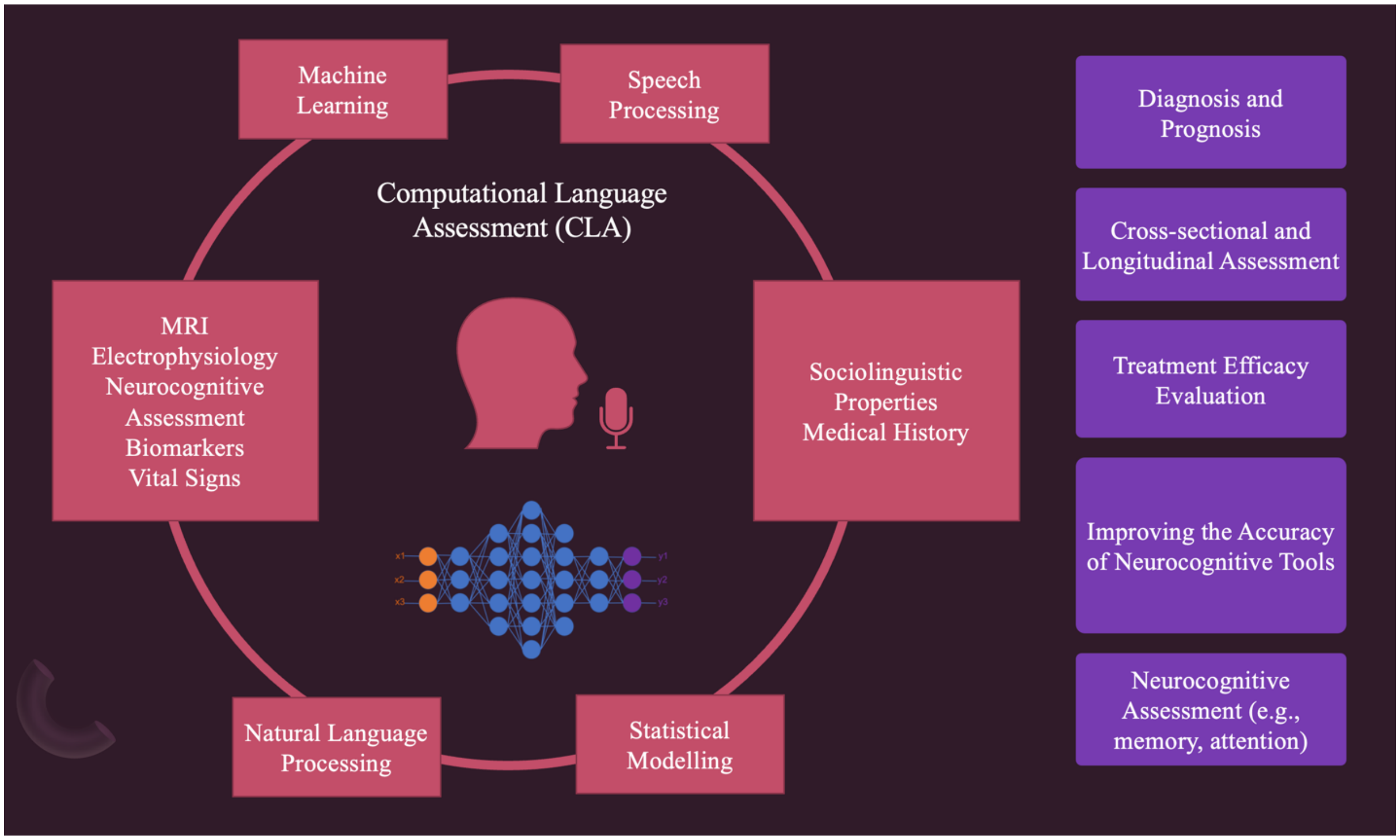

Figure 1 Computational Language Assessment Methodologies and Applications for Evaluating Patients' Speech, Language, and Communication Functioning. The methods include machine learning, speech processing, natural language processing, and statistical modeling. Additional

measures (e.g., imaging, biomarkers, sociolinguistic properties) can be included during machine learning modeling for CLA applications.

### 4.1 Machine Learning Models

Machine learning is a diverse field that encompasses a wide range of models, including Support Vector Machines (SVM), Decision Trees (DT), Random Forests (RF), Principal Components Analysis (PCA), and k-means, to name just a few. However, in recent years, Deep Neural Networks (DNNs) have emerged as an extraordinarily successful approach to solving complex problems. They are widely applied in various areas, including self-driving cars and robotics. Research on neural networks from its early beginnings aimed to model cognitive representation, including language (e.g., McCulloch & Pitts, 1943; Rosenblatt, 1958; Turing, 1936; Turing, 1950). McCulloch and Pitts (1943) simple neural network could compute logical operations**.** However, learning was made possible a decade later by networks proposed by Rosenblatt (1958) and Hebb (1949) and through the formulation of backpropagation, which enabled learning in multi-layered networks in the late 1980s (Rumelhart et al., 1985; Rumelhart et al., 1988; Rumelhart et al., 1986). The current deep neural network approaches, the availability of data, and improved hardware enable more complex high-performance learning (Hassabis et al., 2017; Lecun et al., 2015; Schmidhuber, 2015). Deep neural networks are currently employed for different machine learning tasks, such as supervised, unsupervised, and reinforcement learning (see Figure 2 and Table 1).

Finally, an essential aspect of learning in models, which is often underemphasized, is that it is constantly evolving; thus, machine learning models are always in a state of continuous development, which allows CLA models to improve over time with the addition of more data, better data, and the development of algorithms and methods, which is often counterintuitive to clinicians and researchers that have been using either the same tools (e.g., the same manual neurocognitive batteries and tests) or approaches, such as statistical models for hypothesis testing, where the models are approached in general as static evaluations of several data in a particular point in time.

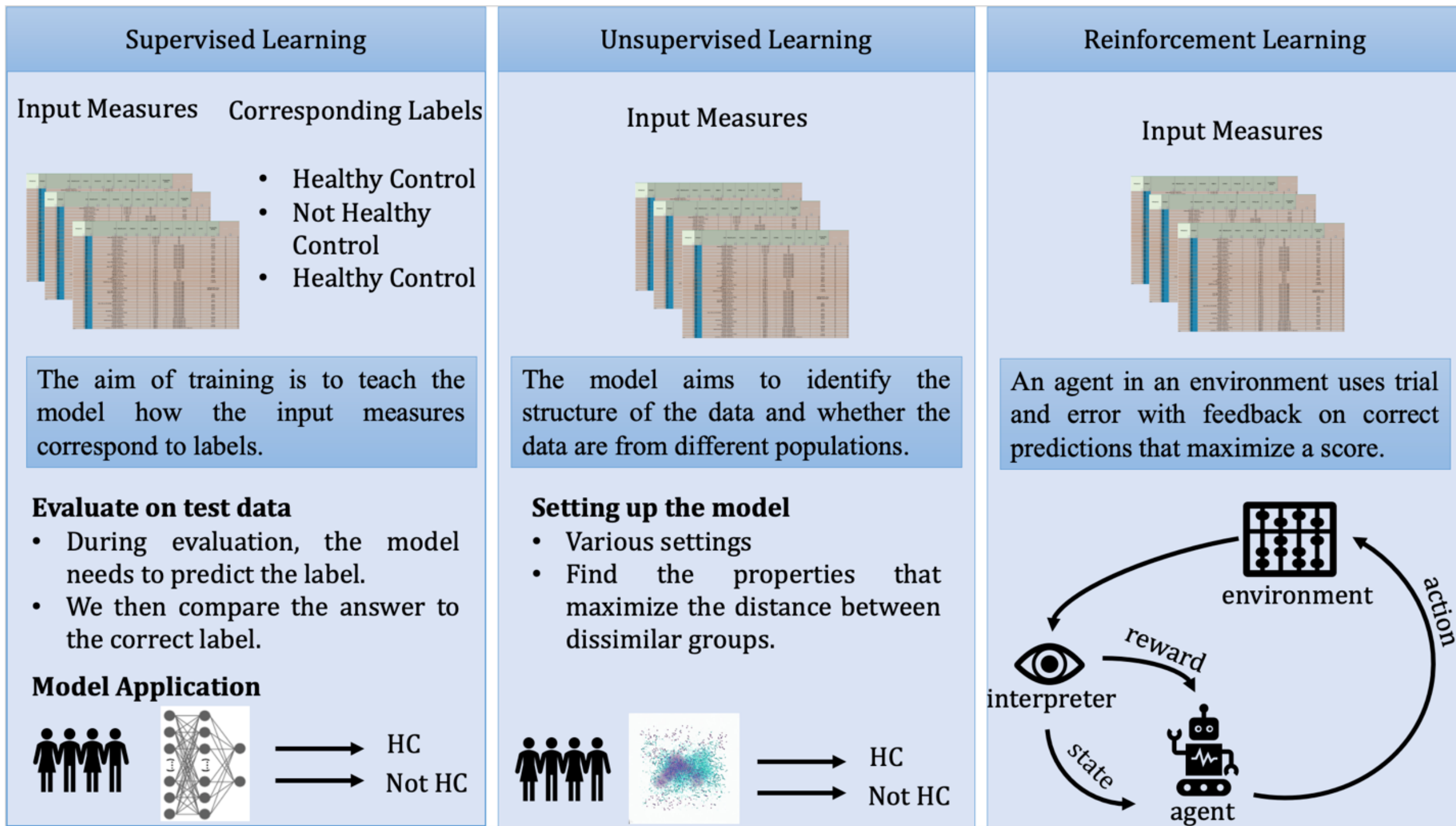

**Figure 2.** Supervised learning aims to teach how the features correspond to the label so that the model can predict unknown data. Unsupervised learning is a method where the model identifies structure in the input data. Semi-unsupervised learning may involve a combination of unsupervised and supervised learning. Reinforcement learning is a method of learning incrementally by adjusting the model. Learned behavior is accompanied by positive feedback or reward, whereas behavior to be avoided is accompanied by negative feedback.

Table 1 Machine Learning Approaches.

| **Term** | Definition |
|---|---|
| **Supervised Learning** | Supervised learning is a learning method where the input variables- a set of features- and the output variable- the label that describes the features- are provided simultaneously. Supervising training aims to teach how the input features correspond to the label clinicians or other experts provided. After training the model, it is evaluated on how well it can predict unknown data. |
| **Unsupervised Learning** | Unsupervised learning is a method where the model identifies structure in the input data without any provided features. Semi-unsupervised learning may involve a combination of unsupervised and supervised learning. |
| **Reinforcement Learning** | Reinforcement learning is a method of learning incrementally by adjusting the network. Learned behavior is accompanied by positive feedback or reward, whereas behavior to be avoided is accompanied by negative feedback. Positive and negative feedback depending on |

| | the application; for example, a network that learns to play a game, positive feedback is earning points and negative feedback is losing points. |
|---|---|

Deep neural network approaches are currently applied successfully in several language-related tasks for analyzing text, audio, and video (Baroni et al., 2017; Lazaridou et al., 2016). For example, recurrent neural networks, such as *long short-term memory units*, simple *feedforward neural networks,* and *convolutional neural networks* (Bengio et al., 2012; Lecun et al., 2015; Schmidhuber, 2015) have been employed to elicit information from texts and for the linguistic analysis of large language corpora. Also, neural networks capture semantic relationships—word embeddings—by identifying relationships between words (Le & Mikolov, 2014; Mikolov et al., 2013; Pennington et al., 2014; Sutskever et al., 2014; Turian et al., 2010).

To address the different tasks and applications, such as classification tasks, text generation, automatic translation, and text summarization, the neural networks are combined in complex DNN architectures, such as those listed in Table 2.

Table 2 Main Neural Network Architectures.

| Term | Definition |
|---|---|
| Feedforward neural network | A network architecture with linear connections between the units that make up the network. |
| Recurrent Neural Networks and Long short-term memory | A type of recurrent neural network architecture that has been applied extensively in speech and text as they can model temporal relations between events. A disadvantage of these networks is that they are slow. |
| Convolutional neural network | A neural network that is inspired by the visual cortex and employs filters to identify significant patterns. It has been used extensively in image recognition. |
| Generative Models | A generative model, unlike a discriminative model, is a model that generates both observed and target values for a phenomenon, given several hidden parameters. |
| Transformers | Transformers are like recurrent neural networks (RNNs) and long short-term memory (LSTM) networks in that they map sequences of words but differ in how they handle long-range dependencies and context. RNNs and LSTMs process sequences sequentially, which makes them slow and prone to forgetting or ignoring distant information. On the other hand, transformers use self-attention mechanisms to process the entire sequence in parallel, allowing them to capture long-range dependencies and context more effectively. |
| Large Language Models | Large language models (LLMs) are based on transformers but are trained on massive amounts of text data using unsupervised learning. This enables them to learn complex patterns and generate realistic and coherent text. Some examples of LLM.s are BERT, developed by Google, and GPT-3, developed by OpenAI. |

## 4.2 Natural Language Processing

Natural Language Processing (NLP) methods, in combination with Deep Neural Networks, are profoundly transforming the analysis of the human language (Asgari et al., 2017; Calzà et al., 2021; Fraser, Lundholm Fors, Eckerström, et al., 2019; Themistocleous, Eckerström, et al., 2018; Tóth et al., 2018). NLP involves techniques for analyzing, studying, processing, and developing human language applications. Thus, CLA employs NLP to provide fast, efficient, and reliable quantification of speech, language, and communication in combination with machine learning and speech analysis. Using these techniques, studying grammar becomes feasible for long texts, with accuracy, ease, and improved transparency, which is impossible when different humans perform the analysis.

### 4.2.1 Lexical and Morphophonological Analysis

The potential of deep neural networks has been known since the 1980s. However, the acceleration of the field in the past three years is unparalleled, making deep neural networks the leading paradigm in machine learning for language analysis. Lexical and morphophonological features characterize the speech of patients with speech and language impairment and healthy controls. By answering this question, we seek to provide a phonological (Vitevitch & Storkel, 2013) (e.g., measures of prosody and vowel quality) (Themistocleous, Eckerström, et al., 2020b), morphological, lexical, syntactic, and semantic biomarkers that distinguish patients from healthy controls. Researchers can employ NLP methods (e.g., tokenization, tagging, and syntactic parsing) and machine learning to elicit novel grammatical measures and evaluate existing ones from the automated transcriptions of speech productions and writings. Researchers can analyze the texts for grammatical measures and evaluate the role of related measures, e.g.,

- *Utterance length*
- *Phonemes-to-word ratio:* e.g., Do speakers prefer long or short words?
- *Content words:* e.g., Nouns, verbs, adjectives, and adverbs.
- *Function Words:* e.g., Conjunctions, e.g., and, or, and, but Prepositions, e.g., de, in, pre and of; Determines, the and a/an; Pronouns such as he/she/it.
- *Part of Speech Ratio:* e.g., content to function word ratio (Fraser et al., 2014; Saffran et al., 1989; Themistocleous, Ficek, et al., 2021; Themistocleous, Webster, et al., 2020) and nouns to vowels ratio.
- *Morphological and semantic information* about the gender (e.g., male, female, neuter), person (first, second, third), number (e.g., singular, plural), and time (e.g., present, past).

### 4.2.2 Syntactic Analysis

Also, NLP analysis can provide automatic syntactic features to characterize patients with speech and language impairment and healthy controls, for example:

- the calculation of probability estimates of syntactic constituents (e.g., noun phrases and verb phrases),
- syntactic complexity (e.g., dependency depth),
- syntactic roles
- the ratio of coordinated, subordinated, and reduced sentences,
- the number of active and passive sentences,
- counts of dependencies (e.g., average dependencies per sentence)

These are calculated using automated syntactic parsers for English and other languages can identify the syntactic role of textual constituents using automated syntactic analysis. Common parsing approaches are dependency parsing, which finds syntactic relations using a dependency grammar and constituent parsing that follows a constituency grammar and constituent parsing. Both are two common approaches to analyzing the grammatical structure of a sentence in natural language processing (Jurafsky & Martin, 2009).

Dependency parsing involves identifying the grammatical relationships between words in a sentence. It aims to identify the dependency relationships between words, where each word is linked to its "head" word in the sentence. For example, in the sentence "John loves Mary," the word "loves" depends on the subject "John" and the object "Mary." Dependency parsing typically produces a tree-like structure known as a dependency tree, which shows the relationships between the words in the sentence (Jurafsky & Martin, 2009).

Constituent parsing involves identifying the syntactic structure of a sentence by dividing it into smaller sub-phrases or constituents (Jurafsky & Martin, 2009). Constituent parsing typically produces a tree-like structure known as a parse tree, which shows the hierarchical structure of the sentence. For example, in the sentence "The cat chased the mouse," the parse tree would show that "the cat" is the subject, "chased" is the verb, and "the mouse" is the object. Constituent parsing is used to identify the grammatical role of each word in a sentence and to determine the meaning of the sentence.

### 4.2.3 Semantic Analysis

Automatic semantic measures analysis aims to determine which computational semantic features characterize the speech of individuals with speech and language impairment. CLA employs semantic analysis to provide quantified measures of semantic relationships, such as antonyms, synonyms, semantic roles (e.g., agent, recipient, goal, and result), and entity characteristics (e.g., person, location, and company) (Bengio & Heigold, 2014; Fraser, Lundholm Fors, & Kokkinakis, 2019; Lappin & ebrary Inc., 1981; Pennington et al., 2014; Sutskever et al., 2014).

Over the years, several methods have been employed to identify semantic relationships from texts, such as post-processing parsed texts and using regular expressions to annotate locations, such as an address or a phone number. However, more recent semantic models rely on extensive textual corpora to establish semantic relationships often as part of deep neural network architectures, such as recurrent neural networks and pre-trained neural network architectures, like Bidirectional Encoder Representations from Transformer networks (BERT) (Devlin et al., 2018; Vaswani et al., 2017), and more recently GPT4 (Bengio & Heigold, 2014; Fraser, Lundholm Fors, & Kokkinakis, 2019; Lappin & ebrary Inc., 1981; Pennington et al., 2014; Sutskever et al., 2014). Word and sentence embeddings from large language models, such as GPT, can explain semantic relationships, provide semantic distances between words, and have the potential to explain neural activation patterns (Hosseini et al., 2022). The embeddings are calculated using large corpora of texts, and the underlying idea of these models is that words that regularly appear in similar contexts are closer semantically. Language models, such as chatGPT, are an exciting area of research; evaluating the performance of the models is not straightforward, and metrics from these models do not correspond to human cognitive representations, so using them to elicit semantic measures that correspond to human semantic understanding is not straightforward. For example, human and computational semantic representations are formed differently concerning multimodal sensory information and social interaction (Bengio & Heigold, 2014; Pennington et al., 2014).

Word and sentence embeddings and large language models, such as GPT, can be employed to compare semantic productions in patients and healthy controls to measure distance scores between patients' productions and model predictions. For example, word embedding can indicate whether patients produce semantically related or unrelated words in word-finding deficits, like naming.

*Name Entity Recognition (N.E.R.*) is a process of information extraction that can be used to determine how semantic relationships are presented linguistically (Jurafsky & Martin, 2009). For example, Napoleon [Person] was the king of France [Place]. N.E.R. can be helpful in patients with aphasia to reveal semantic impairment concerning semantic domains such as people and places.

#### 4.2.4 Discourse Macrostructure

CLA can elicit information about cohesion and coherence, such as the following:

- *Idea density* is a standardized measure of the number of ideas expressed in the number of words or sentences (Danner et al., 2001); and measures of rhetorical structure (e.g., Elaboration, Attribution, and Joint) (Abdalla et al., 2018). Farias et al. (2012) employed idea density, a computational measure, to measure cognitive decline in the Nun Study, a longitudinal study of cognitive decline.
- *Topics can be elicited using topic classification machine learning models, e.g.*, the number and type of discourse markers, the number of topics, and conversational measures such as background elements ('um' and 'hm').
- *Lexical richness* is an informative measure of lexical impairments that measures, for example, if speakers repeat the exact words or can access a variety of words from the lexicon (e.g., type-*token ratio* (TTR), *Herdan's* C (Herdan, 1955), *Maas's TTR, Mean* *segmental TTR, Moving-Average Type–Token* Ratio (MATTR) (Covington & McFall, 2010), *word variation index, counts on function words, hapax legomena* (i.e., words that appear once in the corpus), and *n-grams,* which are sequences of n (2, 3, or more) words that occur in a text and can be employed to identify the speaker characteristics in a text.
- *Sentiment analysis,* lexical and semantic analysis quantify subjective information from texts to analyze the emotional tone and can provide insights into the attitudes, such as the speaker's *stance,* and positive or negative emotions associated with pathology.

Another significant development is the ability to combine language and video. This is an active area of research. Visual and textual alignment techniques (Arandjelovic & Zisserman, 2017; Aytar et al., 2016; Harwath et al., 2016; Ioffe & Szegedy, 2015; Owens et al., 2015) enable researchers to align events in videos to texts (Karpathy & Fei-Fei, 2015). These can allow novel approaches to discourse assessment by aligning this to the environment. An example was provided by Pusiol et al. (2014), who modeled the joint attention of a child and a caregiver toward an object or location. They analyzed video and texts to identify the signals of joined attention using an unsupervised extraction technique of joined attention episodes from video, which "could give hints regarding robust cues that children might use in addition to, or even instead of, gaze." Their study assumed that joined attention is part of the overall language interaction and a necessary part of language learning.

### 4.3 Acoustic Analysis

Acoustic Analysis can provide automatic measures of the acoustic characteristics, namely segmental (vowels and consonants), prosodic (e.g., intonation, pauses, segmental lengthening), and voice quality production that can serve as diagnostic biomarkers. Speech analysis provides the tools to distinguish identity characteristics of speech production in individuals, which include information about their gender, dialect, emotional situation, physiological condition, cognitive and linguistic functioning (Anastasi et al., 2017a, 2017b; G. Angelopoulou et al., 2018; Aristodemou et al., 2015; Bernardy & Themistocleous, 2017; Themistocleous, 2014, 2016a, 2016b, 2017a, 2017b, 2017c, 2017d, 2019; Themistocleous, Eckerström, et al., 2020a, 2020b; Themistocleous et al., 2019; Themistocleous, Fyndanis, et al., 2021; Themistocleous et al., 2022; Themistocleous & Kokkinakis, 2019; Themistocleous & Logotheti, 2016; Themistocleous et al., 2016a, 2016b; Themistocleous, Webster, et al., 2021).

As speech articulation gets impaired by neurodegeneration, for example, in patients with apraxia of speech, speech analysis can provide diagnostic biomarkers of dementia and distinguish patients with language communication disorders from healthy individuals. These can include prosodic and segmental measures (e.g., fundamental frequency and vowel formants), pause duration (Lopez-de-Ipina et al., 2018; López-de-Ipiña et al., 2015), the number of pauses and filled pauses (e.g., um and hm), and phonological measures (sound deletions, insertions, transpositions), variations in the quality of speech sounds (e.g., vowels and consonants), intonation errors, voice quality impairments (Themistocleous, Eckerström, et al., 2018; Themistocleous, Ficek, et al., 2021).

- *Vowel formants:* the first five formant frequencies (F1...F5).
- *Formant dynamics:* measurements of F1...F5 formant frequencies in steps of 5 from the onset of the vowel (time = 1) to the offset of the vowel (time 100): i.e., 1, 5, 10 . . . 100
- *Vowel Duration*: the duration of vowels.
- *Pause duration* (Mack et al. 2015)
- *Intonation:* Fundamental frequency (F0) and related measures, such as the mean F0.
- *Speech rate and fluency.*
- *Voice Quality:* Harmonic and spectral amplitudes measures of voice quality

Also, measures without physical meaning can characterize patients and healthy controls, such as *i*-vectors and spectral moments (Dehak et al., 2011; DeMarco & Cox, 2013; Glembek et al., 2011; Hautamäki et al., 2013; Jiang et al., 2014; Lee et al., 2014; Scheffer et al., 2011; Themistocleous, 2014, 2016a, 2016b, 2017a, 2017b, 2017c; Themistocleous et al., 2022; Themistocleous et al., 2016b).

The acoustic analysis of speech productions provides explanatory measures that enable the identification of the speech profile of an individual, identifying the speaker's identity and revealing information about that person's dialect, emotional state, and physiological condition (Foulkes et al., 2010; Preston & Niedzielski, 2010; Thomas, 2013). For example, we and others have shown that acoustic analysis of *vowel formants,* namely the peak frequencies in vowel spectra, identify the vowel and its properties, such as whether a vowel is front or back, low or high, and can assess minute differences between populations and the accents or dialect of the speaker (Themistocleous, 2017a, 2017c). Also, acoustic analysis can identify the dialects of speakers from the fricative (Themistocleous, 2017b; Themistocleous et al., 2016b), sonorant (Themistocleous, 2019; Themistocleous et al., 2022), and stop consonants (Themistocleous, 2016a). These differences are vital in revealing differences between patients with articulatory impairments (e.g., patients with nfvPPA/apraxia of speech) (Georgiou & Themistocleous, 2020; Themistocleous, 2017a, 2017d).

For example, Themistocleous, Eckerström, et al. (2020b) employed signal processing techniques and showed that patients with MCI differ significantly from healthy controls concerning voice quality measures (i.e., the amplitude of the first harmonic and the amplitude of the third formant (H1-A3); cepstral peak prominence (a measure of dysphonia); the center of gravity indicating the voice's Mean Energy Concentration, and Shimmer (dB) (i.e., the variability of the amplitude from peak-to-peak (local maxima); and articulation rate/averaged speaking time. Also, using acoustic measurements, we could determine the effectiveness of tDCS treatment vs. sham on patients with P.P.A. speech productions (Themistocleous, Webster, et al., 2021). A subsequent study showed that Swedish patients with MCI differ significantly from healthy controls in prosody, voice quality, and fluency (Themistocleous, Eckerström, et al., 2020b). Acoustic analysis of speech production can determine the speech profile of individuals with dementia and provide early acoustic markers that can predict their pathology (Themistocleous, 2016a, 2017a, 2017d; Themistocleous, Eckerström, et al., 2020b; Themistocleous et al., 2022; Themistocleous,

Kokkinakis, et al., 2018). These studies using acoustic analysis also demonstrated that a detailed characterization of the speech profile of individuals with dementia is pressing as it can provide essential and largely unexplored information for assessing speech impairments (e.g., in vowels and consonants, voice quality, speech fluency, rhythm, and prosody).

Acoustic analysis has been employed to analyze several types of speech production tasks, such as diadochokinetic evaluation (DDK) of speech productions, which involves the rapid repetitions of syllables, such as /pa/-/ta/-/ka/, word repetition tasks, and connected speech productions. Researchers have employed various measures such as the Mel frequency cepstral coefficients (MFCCs), measures of the fundamental frequency, such as the mean squared error (M.S.E.) measured relative to the regression curve, and the regression coefficient of the $F_0$ contour within a frame, the mean value of $F_0$, minimum and the maximum of $F_0$, its value in the onset and offset, its temporal variation (jitter), its variation in amplitude (shimmer), and others (Moro-Velazquez et al., 2019; Orozco-Arroyave et al., 2016). For example, Tsanas et al. (2009) employed acoustic dysphonia measures. The Recurrence Period Density Entropy (RPDE), Detrended Fluctuation Analysis (D.F.A.), and Pitch Period Entropy (P.P.E.) have been proposed for the feasibility of identifying patients with P.D. as telemonitoring markers (Tsanas et al., 2009) and for classifying these patients from healthy controls (Tsanas et al., 2012).

## 5 Using CLA for informing diagnosis, prognosis, and treatment efficacy

CLA integrates well with the existing neurocognitive assessment and has the potential to support diagnosis and provide biomarkers and scores in cross-sectional and longitudinal studies, including monitoring a patient's language functioning over time, assessing treatment efficacy, and improving existing neurocognitive evaluations. These studies are discussed in this section.

### 5.1 (Differential) Diagnosis and Prognosis

During life, the biopsy is the best determiner of dementia (along with postmortem histologic examination). Nevertheless, a biopsy is rarely employed for early dementia diagnosis because of its high-risk ratio. Thus, our a need for diagnostic markers and biomarkers for diagnosis. Recent studies have employed speech acoustics to identify individuals with dementia from healthy controls and provide classification models for diagnosis or subtyping (König et al., 2018; Meilan et al., 2018; Themistocleous, Eckerström, et al., 2018; Themistocleous & Kokkinakis, 2019; Tóth et al., 2018). These studies employed measures from (i) segments (i.e., vowels and consonants); (ii) prosody; and (iii) voice quality and speech fluency and showed that they could provide reliable diagnostic markers. In addition, other studies have employed lexical and morphosyntactic to distinguish speakers with speech and language pathology and other conditions (see Figure 2).

TensorFlow (2015), Keras (2015), and PyTorch (2016) sparked an era of enthusiasm about neural networks and their potential in neurology and neuroscience. Among the first studies to utilize these tools for CLA was Themistocleous, Eckerström, et al. (2018), who developed deep neural network models that classify individuals with MCI from Healthy Controls. First, they recorded patients and healthy individuals during the cookie-theft picture description task. Subsequently, the recordings were automatically transcribed and segmented. Next, they measured pauses, vowel duration, and vowel formants (i.e., the frequencies that distinguish one vowel from another). Their model classified individuals with MCI and healthy controls with high classification accuracy (M = 83%) from just one minute of conversation. In addition, their model contributed to the early diagnosis of cognitive decline and showed that acoustic markers facilitate the identification of patients with MCI.

Other studies demonstrated that lexical, morphosyntactic, and textual features provide language biomarkers and distinguish patients with early-onset dementia from healthy controls. Language markers can also be employed for *differential diagnosis* determined within a group of patients. For example, Fraser et al. (2014) observe that differential diagnosis is hard to identify using standardized tests, especially in the initial stages of the condition. In contrast, connected speech has the potential to differentiate subtypes of patients. In their study, they employed natural language processing to elicit textual measures. More specifically, they studied the number of words. They provided measures of syntactic complexity estimated using the Stanford parser. One is the Yngve depth, which quantifies left-branching vs. right-branching phrases calculated from a syntactic tree (Yngve, 1960) and measurements of Parts of Speech (P.O.S.), such as the number of nouns, verbs, and prepositions. These measures distinguished individuals with semantic dementia (S.D.), progressive nonfluent aphasia (PNFA), and healthy controls. They developed three models: the Naïve Bayes, logistic regression, and Support Vector Machines.

Themistocleous, Ficek, et al. (2021) used CLA to provide a classification that follows the consensus criteria for PPA classification proposed by Gorno-Tempini et al. (2011). They employed acoustic, such as the fundamental frequency, pause duration, and morphosyntactic features, and trained Random Forests, Support Vector Machines, and Decision Trees. The latter, a feedforward neural network, outperformed the other machine learning models and expert clinicians' classifications with 80% classification accuracy. Their study showed that 90% of patients with nfvPPA and 95% with lvPPA were identified correctly. Others have developed Naïve Bayes models for dementia diagnosis (Garrard et al., 2014). Naïve Bayes, often with textual features generation (e.g., bag-of-words), were employed to develop good classification models for versatile applications, such as authorship identification and spam detection.

## 5.2 Cross-sectional and longitudinal assessments of language functioning

In clinical practice, manual scoring of language performance is extraordinarily time-consuming and challenging and requires substantial expertise. CLA provides a fast, objective, and quantified score of speech and language performance and reduces trivial tasks, enabling clinicians to focus on patients rather than trivial tasks. CLA has been employed to assess speech fluency, prosody, morphosyntax, lexical, and semantic aspects in patients' speech.

In a recent study, researchers scored *speech fluency* and prosody and showed that these speech features differed significantly between individuals with Mild Cognitive Impairment and Healthy Controls. For example, Themistocleous, Eckerström, et al. (2020b) employed signal processing to study patients' voice quality and speech fluency, showing that voice quality in patients with MCI differed significantly from H.C. Namely, the amplitude of the first harmonic and the amplitude of the third formant (H1-A3); cepstral peak prominence (a measure of dysphonia); the center of gravity indicating the voice's Mean Energy Concentration, and Shimmer (dB) (i.e., the variability of the amplitude from peak-to-peak (local maxima); and articulation rate/averaged speaking time[1].

The evaluation of spelling is a complex, challenging, and time-consuming process. It relies on comparing letter-to-letter words spelled by the patients to the target words. Therefore, Themistocleous, Neophytou, et al. (2020) developed a spelling distance algorithm that automatically compares the inversions, insertions, deletions, and transpositions required to make the target word and the response the same (Neophytou et al., 2018; Themistocleous, Neophytou, et al., 2020).

NLP, signal processing, and machine learning can provide automated measures of language that function as *language biomarkers of dementia.* These involve transcribing a speech and analyzing transcripts using NLP to provide automated part-of-speech (POS) tagging and syntactic parsing. For example, Themistocleous, Webster, et al. (2020) analyzed connected speech productions from 52 individuals with PPA using a morphological tagger. They showed differences in POS production in patients with nfvPPA, lvPPA, and svPPA. This NLP algorithm automatically provides the part of speech category for all words individuals produce (Bird et al., 2009). From the tagged corpus, they measured both content words (e.g., nouns, verbs, adjectives, adverbs) and function words (conjunctions, e.g., and, or, and but; prepositions, e.g., in, and of; determiners, e.g., the a/an, both; pronouns, e.g., he/she/it and wh-pronouns, e.g., what, who, whom; modal verbs, e.g., can, should, will; possessive ending (' s), adverbial particles, e.g., about, off, up; infinitival to, as in to do). Themistocleous, Webster, et al. (2020) showed that the POS patterns of individuals with PPA were both expected and unexpected. It showed that individuals with nfvPPA produced more content words than function words (see top left for the content words and top right for the function words). Individuals with nfvPPA made fewer grammatical words than individuals with lvPPA and svPPA. These studies provide proof of the feasibility that computational tools can be employed to study speech and language. They can form the basis for developing assessment tools for scoring patients' language. By developing novel NLP language models in patients with dementia, researchers analyze language productions from patients with dementia and provide quantified scores of speech, language, and communication symptoms.

### 5.3 Assessment of Treatment Efficacy

CLA applications can implement automated scoring during tasks to provide online feedback about correct or incorrect responses, motivating the patient to try again and self-correct. Thus, these applications can offer treatment and evaluation in any environment. In addition, CLA can support and monitor treatment for patients living in distant areas without access to memory clinics or having mobility issues. Currently, CLA methods provide objective and replicable scores of language performance that take place post hoc after treatment on the recording files, the text transcriptions, and the records that the clinician employed during the neuropsychological or neurocognitive evaluation.

---

[1] Articulation rate is the number of syllables divided by overall duration excluding pauses and silences whereas average speaking time is a measure of the number of syllables divided by the overall duration including pauses and silences).

Using acoustic analysis, Themistocleous, Webster, et al. (2021) provide an example of evaluations between treatments/time points. To evaluate the effects of transcranial direct current stimulation (tDCS), a non-invasive brain stimulation, on speech fluency in patients with nonfluent Primary Progressive Aphasia (nfvPPA) and apraxia of speech (AOS.) as their primary symptom (nfvPPA/AOS), Themistocleous et al. 2021, extracted acoustic measures from vowels and consonants from speech signals. tDCS improves functional connectivity by modulating neuronal excitability by hyperpolarizing or depolarizing the resting membrane potential of neural cells. In other words, neurostimulation modifies the functional connectivity and the gamma-aminobutyric acid (GABA) concentrations. Themistocleous, Webster, et al. (2021) showed that tDCS with language therapy enables patients to speak faster, reflected in vowels and consonants' duration. Sound duration provides an objective measure of fluency, which measures articulatory performance and shows that tDCS combined with language therapy can improve patients' language performance.

### 5.4 Improvement of existing neurocognitive tools

The accuracy of traditional screening tools can also be improved through CLA. Fraser et al. (2018) augmented an ML model for classification by adding automatically extracted linguistic information from a narrative speech task to MMSE scores to increase MMSE's reliability and evaluation outcomes. As a result, the ML model improved by adding just four linguistic features from POS. In addition, the AUC score (measuring a trade-off between sensitivity and specificity) improved from 0.68 to 0.87 in logistic regression classification; a similar increase of accuracy was found by adding prosodic and fluency measures (Themistocleous, Eckerström, et al., 2020a) employed only acoustic measures from speech fluency and prosody. These studies show that models can corroborate existing tools, especially ones that assess linguistic and non-linguistic cognitive domains. The findings show that CLA can complement clinical evaluation and allow for a timely and continuous assessment of patients. Also, it reduces evaluation time and patient stress through automation and enables time allocation for therapy and rehabilitation.

CLA can also help automate the scoring of neurocognitive assessment tools, such as written language evaluations. That has the advantage of providing quantified measures of spelling accuracy. For example, a method for comparing written speech productions concerning target words was developed by Themistocleous, Neophytou, et al. (2020), who employed a modified Levenshtein distance to measure the spelling performance of patients in words and pseudowords. The Levensthein distance is a measure used in NLP for applications that need to compare two strings of letters. This algorithm was also applied to pseudo-words by converting pseudo-words to the International Phonetic Alphabet (IPA) and comparing phonemes of the target and response word with the Levenshtein distance. The advantage of this method is that it allows an efficient evaluation of spelling performance and improves the accuracy and speed of traditional manual scoring methods.

## 6 Conclusions

Computational Language Assessment offers an unprecedented possibility to analyze quickly and efficiently large amounts of data and thus support diagnosis, prognosis, disease progression, and treatment efficacy. Measures are estimated automatically using signal processing, natural language processing, statistical and probabilistic models, and machine learning models. Data are analyzed with CLA, e.g., texts, sound recordings, or neurocognitive tasks. Below, we summarize the significance of CLA methods in language research in neurodegenerative disorders.

1. *Reproducibility and standardization of measurement, terminology, and evaluation.* Standardization promotes the precision of measures across studies, patients, and time points. For example, a trained ML model, such as Random Forests, Neural Networks, or applications like morphosyntactic taggers and parsers, can be reproduced given the same code, data, and training process. CLA provides relatively consistent outcomes for the same input compared to humans, who can differ in their analysis depending on their expertise, the theoretical paradigms they adhere to, the available time, and other factors and limitations that are challenging to predict (Stark et al., 2022; Stark Brielle et al., 2020). Similarly, the acoustic analysis provides measures of speech production that objectively and precisely define the quality of speech sounds (e.g., vowels and consonants) and grants an unbiased estimation of speech performance (Themistocleous et al., 2022; Themistocleous, Webster, et al., 2021). CLA evaluation metrics are standardized measures based on shared principles for distinct models, such as supervised classification, unsupervised classification, binary or multifactorial classification, balanced or unbalanced designs, and reinforcement learning. Also, each machine learning model requires conventional reporting about the architecture and other hyperparameters machine learning models, such as artificial neural networks (e.g., feedforward neural networks, recurrent neural networks, and convolutional neural networks), support vector machines, decision trees, and random forests.
2. *Open-access research.* Publications that provide CLA methods and outputs are freely distributed online, and the code is available for replication and review. This research philosophy promotes good scientific practices ensuring improvement of research techniques and methods of science. Researchers can read code, spot bugs, and trace the thinking process that underlies the development of the model. Also, CLA applications commonly rely on free and open-source languages, such as Python, R, and Julia, which allows the evaluation of their source code. These are stored repositories like GitHub, GitLab, and the Open Science Framework (OSF).
3. *Objective disease and treatment monitoring and increase of our scientific understanding of disease.* The automatic measurements provide quantified results of acoustic and linguistic production and perception and do not rely on the subjective and challenging to quantify perception by humans.

## Challenges and future directions

Since CLA is a developing field, research must solve extraordinary problems. It is essential to determine the selection of data (number of data, the type of data, whether the data are labeled or unlabeled) and the definition of model architectures (e.g., machine learning models). Although no single model can address all aspects of language production, perception, and representation, we must determine the best models for the task. Moreover, as scaling increases and multimodality is introduced into the model, the capabilities of these models in domains other than language increase. These models are well suited for CLA application. They enable multimodal information (e.g., language, MRI, EEG, EcOG, neurocognitive tests, and social factors) and interaction with other cognitive domains such as memory and attention (Diogo et al., 2022; Hosseini et al., 2022). For example, CLA methods easily quantify and address several critical issues, i.e., the effects of brain atrophy on language performance, and provide indications about brain atrophy from language measures. In particular, typical functional and structural magnetic resonance imaging (MRI), anatomical connectivity and functional connectivity measures *(Sporns, 2011)*, and electrophysiological measures, such as EcOG and EEG, can be correlated with automated measures of language functioning to provide biomarkers of dementia and facilitate diagnosis and prognosis *(Wilson et al., 2010)*. Thus, computational CLA models can be employed in interfaces such as brain-to-speech, allowing individuals to speak through neuroprosthesis (Bhaya-Grossman & Chang, 2022; Metzger et al., 2022). Therefore, studies may require interfaces with other domains, such as MRI imaging or neurophysiological measures.

To conclude, CLA can provide a neuro-cognitive evaluation of speech, language, and communication in the elderly and individuals at elevated risk for dementia; they can facilitate the development of computational models for diagnosing, prognosis, and evaluating progression and treatment efficacy in at-risk and language-impaired populations and allow easier extensibility to assess patients from a wide range of languages. In addition, CLA-based AI models can inform theory on the relationship between language symptoms and their neural bases and significantly advance our ability to optimize the prevention and treatment of elderly individuals with communication disorders, allowing them to age gracefully with social engagement. Ultimately, CLA methods for neurocognitive assessments can facilitate a move towards a more inclusive, equitable, and democratic approach to science, where everyone has access to high-quality assessments and care regardless of their background, financial status, and identity.

**Competing Interests:** The authors have no competing interests or other interests that might be perceived to influence the results and discussion reported in this paper. The author has received no funding for this project from other organization.